\documentclass[twoside,leqno,twocolumn]{article}
\usepackage{ltexpprt}
\usepackage{booktabs} 
\usepackage{float}
\usepackage{pgfplots}
\usepgfplotslibrary{colorbrewer}
\usepackage{amsfonts}
\usepackage{amsmath}
\usepackage[export]{adjustbox}

\usepackage{tikz}
\usetikzlibrary{positioning}
\usetikzlibrary{shapes.geometric}

\usepackage[font=bf]{caption}
\usepackage{url}
\usepackage{enumitem}

\usepackage{hhline}
\usepackage{multirow}
\usepackage{booktabs}
\usepackage{wrapfig}
\usepackage{xcolor}
\usepackage{xspace}

\usepackage{hhline}
\usepackage{multirow}
\usepackage{booktabs}
\usepackage{algorithm}
\usepackage{algpseudocode}
\usepackage[algo2e,linesnumbered]{algorithm2e} 
\renewcommand{\@algocf@capt@plain}{above}

\usepackage{subfigure}
\usepgfplotslibrary{colorbrewer}
\usetikzlibrary{patterns}

\newcommand{\set}[1]{\left\{#1\right\}}
\newcommand{\real}{\mathbb{R}}
\newcommand{\abs}[1]{\ensuremath{|#1|}\xspace}
\newcommand{\defines}{\ensuremath{\leftarrow}\xspace}

\newcommand{\dimension}{D}
\newcommand{\distfunc}{d}

\newcommand{\branchfactor}{K}
\newcommand{\treemaxdepth}{\ensuremath{d_{\text{max}}}\xspace}

\newcommand{\vect}[1]{\textbf{#1}}
\newcommand{\labelrepr}{\vect{v}}
\newcommand{\linearsep}{\vect{w}}
\newcommand{\centerrepr}{\textbf{c}}
\newcommand{\labelset}{S}

\newcommand{\trainsize}{N}

\newcommand{\bonsai}{\texttt{Bonsai}\xspace}
\newcommand{\parabel}{\texttt{Parabel}\xspace}

\newcommand{\rootnode}{\ensuremath{r}\xspace}

\newcommand{\childnode}{\ensuremath{\node_{k}}\xspace}
\newcommand{\childrennodes}{\ensuremath{\set{\node_1, \ldots, \node_\branchfactor}\xspace}}

\newcommand{\onevsall}{One-vs-All\xspace}

\newcommand{\pwt}[1]{prec\_wt$@$#1}
\newcommand{\nwt}[1]{nDCG\_wt$@$#1}

\newcommand{\pointx}{\ensuremath{\vect{x}}\xspace}
\newcommand{\pointy}{\ensuremath{\vect{y}}\xspace}
\newcommand{\leaf}{\ensuremath{e}\xspace}
\newcommand{\ancestor}{\ensuremath{\mathcal{A}}\xspace}
\newcommand{\node}{\ensuremath{n}\xspace}
\newcommand{\parent}[1]{\ensuremath{p(#1)}\xspace}
\newcommand{\event}[1]{\ensuremath{z_{#1}}\xspace}
\newcommand{\proba}{\ensuremath{\text{Pr}}\xspace}
\newcommand{\nodeproba}{\ensuremath{\proba(\event{n} = 1 \mid \event{\parent{n}}=1,\pointx)}\xspace}
\newcommand{\leafproba}{\ensuremath{\proba(\pointy_{\ell} = 1 \mid \event{\leaf}=1,\pointx)}\xspace}
\newcommand{\predictproba}{\ensuremath{\proba(\pointy_{\ell}=1 \mid \pointx)}\xspace}

\newcommand{\grow}{\textsc{grow}}

\newcommand{\solveonevsall}{\textsc{one-vs-all}}

\newcommand{\instances}{\ensuremath{\mathcal{I}}\xspace}
\newcommand{\labelparts}{\ensuremath{S_1, \ldots, S_\branchfactor}\xspace}

\newcommand{\children}{\ensuremath{\mathcal{C}}\xspace}
\newcommand{\labels}{\ensuremath{\mathcal{L}}\xspace}
\newcommand{\depth}{\ensuremath{d}\xspace}%

\newcommand{\patk}{\ensuremath{prec@k}\xspace}
\newcommand{\ndcgatk}{\ensuremath{nDCG@k}\xspace}%

\newcommand{\bonsaii}{\texttt{Bonsai-i}\xspace}
\newcommand{\bonsaio}{\texttt{Bonsai-o}\xspace}
\newcommand{\bonsaiio}{\texttt{Bonsai-io}\xspace}

\begin{document}
\title{Bonsai - Diverse and Shallow Trees for Extreme Multi-label Classification}


\author{Sujay Khandagale$^1$, Han Xiao$^2$ and Rohit Babbar$^2$
}
\date{
$^1$Indian Institute of Technology Mandi, India \\
$^2$Aalto University, Helsinki, Finland\\ [2ex]%
}

\maketitle              

\begin{abstract}

Extreme multi-label classification (\textbf{XMC}) refers to supervised multi-label learning involving hundreds of thousand or even millions of labels. 
In this paper, we develop a suite of algorithms, called \texttt{Bonsai}, which generalizes the notion of label representation in XMC, and partitions the labels in the representation space to learn shallow trees.
We show three concrete realizations of this label representation space including : (i) the input space which is spanned by the input features, (ii) the output space spanned by label vectors based on their co-occurrence with other labels, and (iii) the joint space by combining the input and output representations. 
Furthermore, the constraint-free multi-way partitions learnt iteratively in these spaces lead to shallow trees.

By combining the effect of shallow trees and generalized label representation, \texttt{Bonsai} achieves the best of both worlds - fast training which is comparable to state-of-the-art tree-based methods in XMC, and much better prediction accuracy, particularly on tail-labels.
On a benchmark Amazon-3M dataset with 3 million labels, \bonsai outperforms a state-of-the-art one-vs-rest method in terms of prediction accuracy, while being approximately 200 times faster to train. 
The code for \bonsai is available at \url{https://github.com/xmc-aalto/bonsai}.
\end{abstract}


\section{Introduction}
\label{sec:intro}
Extreme Multi-label Classification (\textbf{XMC}) refers to supervised learning of a classifier which can automatically label an instance with a small subset of relevant labels from an extremely large set of all possible target labels.
Machine learning problems consisting of hundreds of thousand labels are common in various domains such as product categorization for e-commerce \cite{mcauley2013hidden,shen2011item,Bengio10,Agrawal13}, hash-tag suggestion in social media \cite{denton2015user}, annotating web-scale encyclopedia \cite{partalas2015lshtc}, and image-classification \cite{krizhevsky2012imagenet,deng2010does}.
It has been demonstrated that, the framework of XMC can also be leveraged to effectively address ranking problems arising in bid-phrase suggestion in web-advertising and suggestion of relevant items for recommendation systems \cite{prabhu2014fastxml}.

From the machine learning perspective, building effective extreme classifiers is faced with the \textit{computational} challenge arising due to large number of (i) output labels, (ii) input training instances, and (iii) input features. 
Another important statistical characteristic of the datasets in XMC is that a large fraction of labels are \textit{tail labels}, i.e., those which have very few training instances that belong to them (also referred to as power-law, fat-tailed distribution and Zipf's law). 
Formally, let $N_r$ denote the size of the $r$-th ranked label, when ranked in decreasing order of number of training instances that belong to that label, then :
\begin{equation}
N_r = N_1r^{-\beta}
\label{eq:powerlaw1}
\end{equation} 
\begin{figure}[!h]
\centering
\includegraphics[width=\columnwidth]{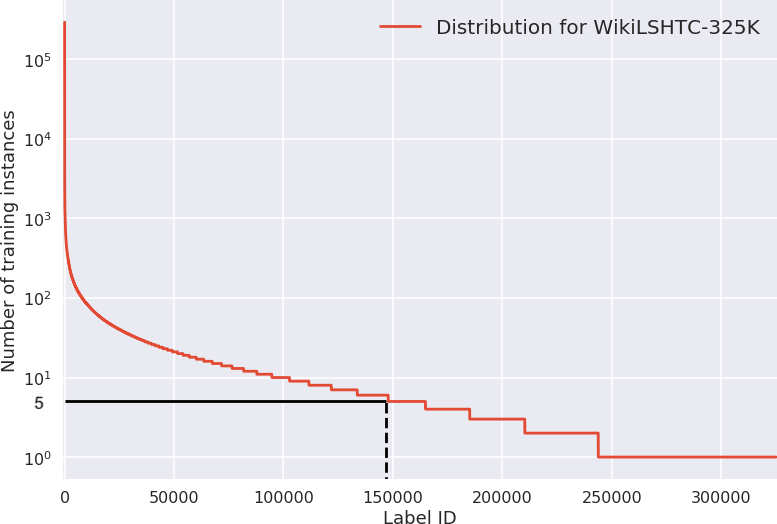}
\caption{
Label frequency in dataset WikiLSHTC-325K shows power-law distribution. 
X-axis shows the label IDs sorted by their frequency in training instances and Y-axis gives the actual frequency (on log-scale). 
Note that more than half of the labels have fewer than 5 training instances.
}
\label{fig:powerlaw}
\end{figure}
where $N_1$ represents the size of the 1-st ranked label and $\beta>0$ denotes the exponent of the power law distribution.
This distribution is shown in Figure \ref{fig:powerlaw} for a benchmark dataset, WikiLSHTC-325K from the XMC repository \cite{repo}.
In this dataset, only $\sim$150,000 out of 325,000 labels have more than 5 training instances in them. 
Tail labels exhibit diversity of the label space, and contain informative content not captured by the head or torso labels.
Indeed, by predicting well the head labels, yet omitting most of the tail labels, an algorithm can achieve high accuracy \cite{wei2018does}.
However, such behavior is not desirable in many real world applications, where fit to power-law distribution has been observed \cite{babbar2014power, Babbar2019}.


\subsection{Related work}
Various works in XMC can be broadly categorized into one of the four strands :
\begin{enumerate}

\item \textit{One-vs-rest} : As the name suggests, these methods learn a classifier per label which distinguishes it from rest of the labels. In terms of prediction accuracy and label diversity, these methods have been shown to be among the best performaning ones for XMC~\cite{dismec,yen2017ppdsparse, Babbar2019}. 
However, due to their reliance on a distributed training framework, it remains challenging to employ them in resource constrained environments.

\item \textit{Tree-based} : Tree-based methods implement a divide-and-conquer paradigm and scale to large label sets in XMC by partitioning the labels space. As a result, these scheme of methods have the computational advantage of enabling faster training and prediction~\cite{prabhu2014fastxml,Jain16,jasinska,majzoubi2019ldsm, wydmuch2018no}. Approaches based on decision trees have also been proposed for multi-label classification and those tailored to XMC regime \cite{joly2019gradient, si2017gradient}. However, tree-based methods suffer from error propagation in the tree cascade as also observed in hierarchical classification \cite{babbar2013flat,babbar2016learning}. As a result, these methods tend to perform particularly worse on metrics which are sensitive for tail-labels~\cite{prabhu2018parabel}.

\item \textit{Label embedding} : Label-embedding approaches assume that, despite large number of labels, the label matrix is effectively low rank and therefore project it to a low-dimensional sub-space.
These approaches have been at the fore-front in multi-label classification for small scale problems with few tens or hundred labels \cite{hsu2009multi, tai2012multilabel, weston2011wsabie, linmulti}. For power-law distributed labels in XMC settings, the crucial assumption made by the embedding-based approaches of a low rank label space breaks down \cite{xurobust, bhatia2015sparse, tagami2017annexml}.
Under this condition, embedding based approaches leads to high prediction error.

\item \textit{Deep learning} : Deeper architectures on top of word-embeddings have also been explored in recent works~\cite{liu2017deep,joulin2017bag,mikolov2013distributed}. However, their performance still remains sub-optimal compared to the methods discussed above which are based on bag-of-words feature representations. This is mainly due to the data scarcity in tail-labels which is substantially below the sample complexity required for deep learning methods to reach their peak performance.

\end{enumerate}

Therefore, a central challenge in XMC is to build classifiers which retain the accuracy of one-vs-rest paradigm while being as efficiently trainable as the tree-based methods.
Recently, there have been efforts for speeding up the training of existing classifiers by better initialization and exploiting the problem structure \cite{fang2019fast, liang2018block, jalan2019accelerating}.
In a similar vein, a recently proposed tree-based method, \parabel~\cite{prabhu2018parabel}, partitions the label space recursively into two child nodes using 2-means clustering.
It also maintains a balance between these two label partitions in terms of number of labels.
Each intermediate node in the resulting binary label-tree is like a meta-label which captures the generic properties of its constituent labels. 
The leaves of the tree consist of the actual labels from the training data. During training and prediction each of these labels is distinguished from other labels under the same parent node through the application of a binary classifier at internal nodes and one-vs-all classifier for the leaf nodes. 
By combination of tree-based partitioning and one-vs-rest classifier, it has been shown to give better performance than previous tree-based methods~\cite{prabhu2014fastxml,Jain16,jasinska} while simultaneously allowing efficient training.

However, in terms of prediction performance, \parabel remains sub-optimal compared to one-vs-rest approaches. 
In addition to error propagation \textit{due to cascading effect of the deep trees}, its performance is particularly worse on tail labels. 
This is the result of two strong constraints in its label partitioning process, (i) each parent node in the tree has only \textit{two} child nodes, and (ii) at each node, the labels are partitioned into \textit{equal sized} parts, such that the number of labels under the two child nodes differ by at most one. 
As a result of the coarseness imposed by the binary partitioning of labels, the tail labels get subsumed by the head labels.

\definecolor{blue}{HTML}{2996B2}
\definecolor{lightblue}{HTML}{7ACBF8}
\definecolor{red}{HTML}{FF5121}
\definecolor{lightred}{HTML}{FFC573}
\definecolor{darkgreen}{HTML}{5ACD20}
\definecolor{green}{HTML}{8EFF2B}
\definecolor{lightgreen}{HTML}{B6FF82}

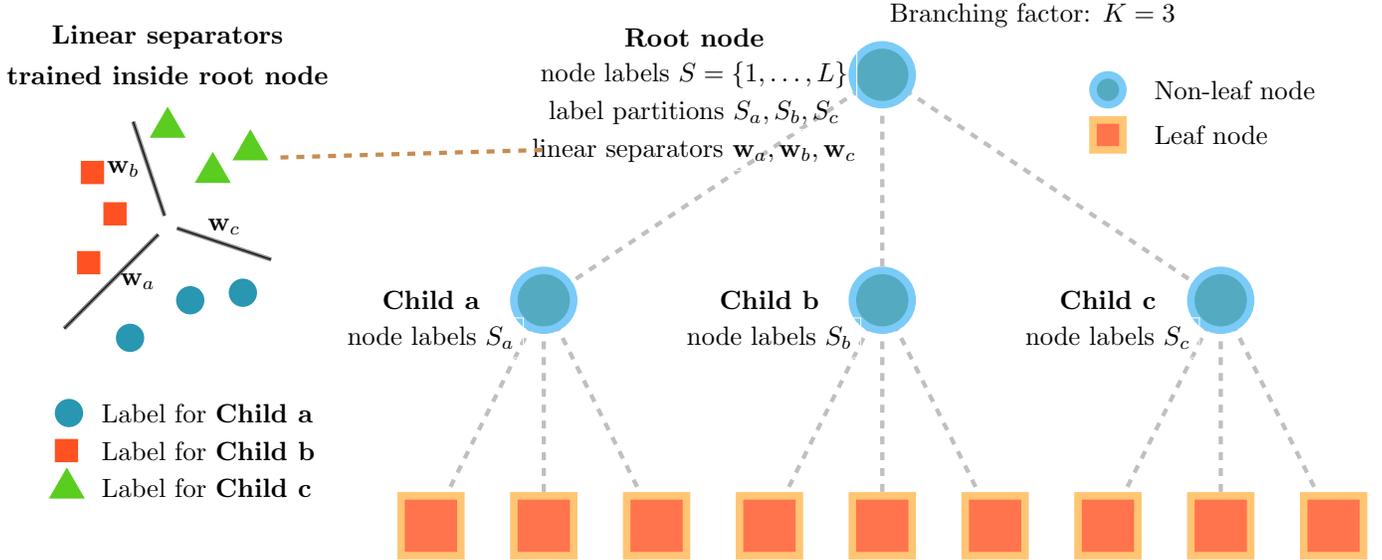
\begin{figure*}[!h]
  \makebox[\textwidth][c]{
    \begin{tikzpicture}[scale=1.0, square/.style={regular polygon,regular polygon sides=4}]

      \tikzstyle{circlenode} = [font=\small, inner sep = 8pt, text=black!70, rounded corners, text centered, circle, line width=0.1cm]
      \tikzstyle{squarenode} = [font=\small, inner sep = 8pt, text=black!70, text centered, square, line width=0.1cm]
      \tikzstyle{leaf} = [draw = lightred, fill = red!80, squarenode]
      \tikzstyle{internal} = [draw = lightblue, fill = blue!80, circlenode]

      \tikzstyle{labelstyle} = [thick]
      \tikzstyle{label1} = [rectangle, draw=red, fill=red, inner sep = 4pt, labelstyle]
      \tikzstyle{label2} = [circle, draw=blue, fill=blue, inner sep = 3.5pt, labelstyle]
      \tikzstyle{label3} = [regular polygon, regular polygon sides=3, draw=darkgreen, fill=darkgreen, labelstyle, inner sep = 2.5pt]

      \tikzstyle{separator} = [ultra thick, draw=gray]
      \tikzstyle{mathbox} = [draw=white, text centered, rectangle, line width=0]
      \tikzstyle{linkstyle} = [dashed, ultra thick, draw=gray!50]
      \tikzstyle{dlink} = [->, linkstyle]
      \tikzstyle{link} = [linkstyle]

      \node at (11,6) (root) {};
      \node at (6.5,3) (child1) {};
      \node at (11,3) (child2) {};
      \node at (15.5,3) (child3) {};
      \node at (5,0) (leaf1) {};
      \node at (6.5,0) (leaf2) {};
      \node at (8,0) (leaf3) {};
      \node at (9.5,0) (leaf4) {};
      \node at (11,0) (leaf5) {};
      \node at (12.5,0) (leaf6) {};
      \node at (14,0) (leaf7) {};
      \node at (15.5,0) (leaf8) {};
      \node at (17,0) (leaf9) {};

      \draw [dlink] (root) -- (child1);
      \draw [dlink] (root) -- (child2);
      \draw [dlink] (root) -- (child3);

      \draw [dlink] (child1) -- (leaf1);
      \draw [dlink] (child1) -- (leaf2);
      \draw [dlink] (child1) -- (leaf3);
      \draw [dlink] (child2) -- (leaf4);
      \draw [dlink] (child2) -- (leaf5);
      \draw [dlink] (child2) -- (leaf6);
      \draw [dlink] (child3) -- (leaf7);
      \draw [dlink] (child3) -- (leaf8);
      \draw [dlink] (child3) -- (leaf9);


      \draw node [internal] at (root) {};
      \draw node[internal] at (child1) {};
      \draw node[internal] at (child2) {};
      \draw node[internal] at (child3) {};

      \draw node[leaf] at (leaf1) {};
      \draw node[leaf] at (leaf2) {};
      \draw node[leaf] at (leaf3) {};
      \draw node[leaf] at (leaf4) {};
      \draw node[leaf] at (leaf5) {};
      \draw node[leaf] at (leaf6) {};
      \draw node[leaf] at (leaf7) {};
      \draw node[leaf] at (leaf8) {};
      \draw node[leaf] at (leaf9) {};

      \draw node [internal, inner sep = 4pt] at (14, 5.8) {};
      \draw node[mathbox, right] at (14.5, 5.8) {Non-leaf node};
      \draw node [leaf, inner sep = 4pt] at (14, 5.2) {};
      \draw node[mathbox, right] at (14.5, 5.2) {Leaf node};

      \draw node[mathbox] at (13, 6.8) {Branching factor: $\branchfactor=3$};

      \node at (1.5, 6.5) (right) {\textbf{Linear separators}};
      \node at (1.5, 6) (right) {\textbf{trained inside root node}};

      \node at (1.5, 4) (sepcenter) {};
      \node at (0, 2.5) (sep1) {};
      \node at (1, 5.5) (sep2) {};
      \node at (3, 3.5) (sep3) {};

      \draw [separator] (sepcenter) -- (sep1);
      \draw (sep1) -- (sepcenter) node [midway, right, text width=2pt] (sep1w) {$\linearsep_a$};

      \draw [separator] (sepcenter) -- (sep2);
      \draw (sep2) -- (sepcenter) node [midway, left] (sep2w) {$\linearsep_b$};

      \draw [separator] (sepcenter) -- (sep3);
      \draw (sep3) -- (sepcenter) node [midway, above] (sep3w) {$\linearsep_c$};

      \node () at (0.5, 4.7) [label1] {};
      \node () at (0.8, 4.15) [label1] {};
      \node () at (0.45, 3.5) [label1] {};
      \node () at (1, 2.5) [label2] {};
      \node () at (1.8, 3) [label2] {};
      \node () at (2.5, 3.1) [label2] {};
      \node () at (2.6, 5.0) [label3] {};
      \node () at (1.5, 5.3) [label3] {};
      \node () at (2.1, 4.7) [label3] {};

      \node () at (0, 1.5) [label2, right] {};
      \node (mathbox) at (0.5, 1.5) [right] {Label for \textbf{Child a}};
      \node () at (0, 1) [label1, right] {};
      \node (mathbox) at (0.5, 1) [right] {Label for \textbf{Child b}};
      \node () at (0, 0.5) [label3, right] {};
      \node (mathbox) at (0.5, 0.5) [right] {Label for \textbf{Child c}};

      \draw node[mathbox] at (8.5, 6.5) {\textbf{Root node}};
      \draw node[mathbox] at (8.5, 6) {node labels $S=\set{1, \ldots, L}$};
      \draw node[mathbox] at (8.5, 5.5) {label partitions $\labelset_a, \labelset_b, \labelset_c$};
      \draw node[mathbox] at (8.5, 5) {linear separators $\linearsep_a, \linearsep_b, \linearsep_c$};

      \draw node[mathbox] at (5, 3) {\textbf{Child a}};
      \draw node[mathbox] at (5, 2.5) {node labels $\labelset_a$};
      \draw node[mathbox] at (9.5, 3) {\textbf{Child b}};
      \draw node[mathbox] at (9.5, 2.5) {node labels $\labelset_b$};
      \draw node[mathbox] at (14, 3) {\textbf{Child c}};
      \draw node[mathbox] at (14, 2.5) {node labels  $\labelset_c$};


      \draw [ultra thick, dashed, draw=brown!90] (3, 4.9) -- (6.5, 5);
    \end{tikzpicture}
  }
  \caption{Illustration of \bonsai architecture. 
    During training, label are partitioned hierarchically, resulting in a tree structure of label partitions. 
    In order to obtain diverse and shallow trees, the branching factor ($\branchfactor$) is set to $\geq 100$ in \bonsai (shown as 3 for better pictorial illustration).
    This is in contrast to \parabel, where it is set to 2, leading to binary label partitions and hence much deeper trees.
    Inside non-leaf nodes, linear classifiers are trained to predict which child nodes to traverse down during prediction. 
    Inside leaf nodes, linear classifiers are trained to predict the actual labels. 
  }
  \label{fig:arch}
\end{figure*}

\subsection{Bonsai overview} 

In this paper, we develop a family of algorithms, called \texttt{Bonsai}. 
At a high level, \bonsai follows a similar paradigm which is common in most tree-based approaches, i.e., label partitioning followed by learning classifiers at the internal nodes. 
However, it has two main features, which distinguish it from state-of-the-art tree based approaches. These are summarized below :

\begin{itemize}
\item \textit{Generalized label representation -} 
In this work, we argue that the notion of representing the labels is quite general, and their exist various meaningful manifestations of the label representation space. As three concrete examples, we show the applicability of the following representations of labels : (i) input space representation as a function of feature vectors (ii) output space representation based on their co-occurrence with other labels, and (iii) a combination of the output and input representations.
In this regard, our work generalizes the view taken in most earlier works, which have represented labels only in the input space such as by representing them as sum of the training instances in which their are active ~\cite{prabhu2018parabel,wydmuch2018no}.
We show that these representations, when combined with shallow trees (described next), surpass existing methods demonstrating the efficacy of the proposed generalization representation.

\item \textit{Shallow trees -} To avoid error propagation in the tree cascade, we propose to construct a shallow tree architecture. This is achieved by enabling (i) a flexible clustering via $K-$means for $K>2$, and (ii) relaxing balancedness constraints in the clustering step. 
Multi-way partitioning initializes diverse sub-groups of labels, and the unconstrained nature maintains the diversity during the entire process.
These are in contrast to tree-based methods which impose such constraints for a balanced tree construction. As we demonstrate in our empirical findings, by relaxing the constraints, \texttt{Bonsai} leads to prediction diversity and significantly better tail-label coverage.
\end{itemize}


By synergizing the effect of a richer label representation and shallow trees, \texttt{Bonsai} achieves the best of both worlds - prediction diversity better than state-of-the-art tree-based methods with comparable training speed, and prediction accuracy at par with one-vs-rest methods.
The code for \bonsai is available at \url{https://github.com/xmc-aalto/bonsai}.

\section{Formal description of Bonsai}
\label{sec:bonsai}





We assume to be given a set of $N$ training points $\set{(\pointx_i, \pointy_i)}_{i=1}^N$ with $\dimension$ dimensional feature vectors $\pointx_i \in \real^\dimension$ and $L$ dimensional label vectors $\pointy_i \in \{0,1\}^L$. 
Without loss of generality, let the set of labels is represented by $\{1,\ldots, \ell, \ldots , L\}$
Our goal is to learn a multi-label classifier in the form of a vector-valued output function $f : \mathbb{R}^D \mapsto \{0,1\}^L$. 
This is typically achieved by minimizing an empirical estimate of $\mathbb{E}_{(\textbf{x},\textbf{y}) \sim \mathcal{D}}[\mathcal{L}(\textbf{W};(\textbf{x},\textbf{y}))]$ where $\mathcal{L}$ is a loss function, and samples $(\textbf{x},\textbf{y})$ are drawn from some underlying distribution $\mathcal{D}$.
The desired parameters $\textbf{W}$ can take one of the myriad of choices.
In the simplest (and yet effective) of setups for XMC such as linear classification, $\textbf{W}$ can be in the form of matrix. 
In other cases, it can be representative of a deeper architecture or a cascade of classifiers in a tree structured topology.
Due to their scalability to extremely large datasets, \bonsai follows a tree-structured partitioning of labels. 

In this section, we next present in detail the two main components of \bonsai : (i) generalized label representation and (ii) shallow trees.

\subsection{Label representation}
\label{sec:label-partitions}

In the extreme classification setting, labels can be represented in various ways. To motivate this, as an analogy in terms of publications and their authors, one can think of labels as authors, the papers they write as their training instances, and multiple co-authors of a paper as the multiple labels. 
Now, one can represent authors (labels) either solely based on the content of the papers they authored (input space representation), or based only on their co-authors (output space representation) or as a combination of the two. 

Formally, let each label $\ell$ be represented by $\eta$-dimensional vector $\textbf{v}_{\ell} \in \mathbb{R}^{\eta}$.
Now, $\textbf{v}_{\ell}$ can be represented as a function \textit{only} of input instances $\set{\pointx_i}_{i=1}^N$, \textit{only} of output instances $\set{\pointy_i}_{i=1}^N$ or as a combination of \textit{both} $\set{(\pointx_i, \pointy_i)}_{i=1}^N$. We now present three concrete realizations of the label representation $\textbf{v}_{\ell}$. We later show that these representations can be seamlessly combined with shallow tree cascade of classifiers, and yield state-of-the-art performance on XMC tasks.

%
%
\begin{itemize}
    \item[a.] \textit{Input space representation of} $\textbf{v}_{\ell}$ - The label representation for label $\ell$ can be arrived at by summing all the training examples for which it is active. Let $\mathcal{V}_i$ be the label representation matrix given by
    \begin{equation}
        \mathcal{V}_i = \vect{Y}^T\vect{X} = \begin{bmatrix}
        \vect{v}_1^T\\
        \vect{v}_2^T\\
        \vdots\\
        \vect{v}_L^T
        \end{bmatrix}_{L \times D} 
    \end{equation}
    \begin{equation*}
        \mathrm{where \quad} \vect{X} = \begin{bmatrix}
        \vect{x}_1^T\\
        \vect{x}_2^T\\
        \vdots\\
        \vect{x}_N^T
        \end{bmatrix}_{N \times D}\mathrm{ ,\quad} \vect{Y} = \begin{bmatrix}
        \vect{y}_1^T\\
        \vect{y}_2^T\\
        \vdots\\
        \vect{y}_N^T
        \end{bmatrix}_{N \times L}.
    \end{equation*}
    We follow the notation that each bold letter such as $\textbf{x}$ is a vector in column format and $\textbf{x}^T$ represents the correponding row vector. 
Hence, each row $\vect{v}_{\ell}$ of matrix $\mathcal{V}_i$ which represents the label $\ell$, is given by the sum of all the training instances for which label $\ell$ is active. 
This can also be represented as, $\vect{v}_{\ell} = \sum_{i=1}^N \vect{y}_{i{\ell}}\vect{x}_i$. 
Note that even though $\vect{v}_{\ell}$ also depends on the label vectors, it is still in the same space as the input instance and has dimensionality $D$.
Furthermore, each $\vect{v}_{\ell}$ can be normalized to unit length in euclidean norm as follows : $\vect{v}_{\ell} := \vect{v}_{\ell}/\|\vect{v}_{\ell}\|_2$.
    \item[b.] \textit{Output space representation of} $\textbf{v}_{\ell}$ - In the multi-label setting, another way to represent the labels is to represent them solely as a function of the degree of their co-occurence with other labels. That is, if two labels co-occur with similar set of labels, then these are bound to be related to each other, and hence should have similar representation. In this case, the label representation matrix $\mathcal{V}_o$ is given by
    \begin{equation}
        \mathcal{V}_o = \vect{Y}^T\vect{Y} = \begin{bmatrix}
        \vect{v}_1^T\\
        \vect{v}_2^T\\
        \vdots\\
        \vect{v}_L^T
        \end{bmatrix}_{L \times L} 
        \mathrm{, where\quad} \vect{Y} = \begin{bmatrix}
        \vect{y}_1^T\\
        \vect{y}_2^T\\
        \vdots\\
        \vect{y}_N^T
        \end{bmatrix}_{N \times L}.
    \end{equation}
    Here $\mathcal{V}_o$ is an $L \times L$ symmetric matrix, where each row $\vect{v}_{\ell}^T$, corresponds to the number of times the label $\ell$ co-occurs with all other labels. 
Hence these label co-occurrence vectors $\vect{v}_{\ell}$ give us another way of representing the label $\ell$. It may be noted that in contrast to the previous case, being an output space representation, the dimensionality of the label vector is same as that of the output space having the same dimensionality, i.e. $\eta=L$.
    \item[c.] \textit{Joint input-output representation of} $\textbf{v}_{\ell}$ - Given the previous input and output space representations of labels, a natural way to extend it is by combining these representations via concatenation. 
    This is achieved as follows, for a training instance $i$ with feature vector $\vect{x}_i$ and corresponding label vector $\vect{y}_i$, let $\vect{z}_i$ be the concatenated vector given by, $\vect{z}_i = [\vect{x}_i \odot \vect{y}_i]$. Then, the joint representation can be computed in the matrix $\mathcal{V}_j$ as follows 
    
    \begin{equation}
        \mathcal{V}_j = \vect{Y}^T\vect{Z} = \begin{bmatrix}
        \vect{v}_1^T\\
        \vect{v}_2^T\\
        \vdots\\
        \vect{v}_L^T
        \end{bmatrix}_{L \times (D+L)}
    \end{equation}
    \begin{equation*}
        \mathrm{where\quad} \vect{Z} = \begin{bmatrix}
        \vect{z}_1^T\\
        \vect{z}_2^T\\
        \vdots\\
        \vect{z}_N^T
        \end{bmatrix}_{N \times (D+L)}\mathrm{ } \vect{Y} = \begin{bmatrix}
        \vect{y}_1^T\\
        \vect{y}_2^T\\
        \vdots\\
        \vect{y}_N^T
        \end{bmatrix}_{N \times L}
    \end{equation*}
    Here each row $\vec{v}_{\ell}$ of the label representation matrix $\mathcal{V}_j$ which is the label representation in the joint space, is therefore a concatenation of representations obtained from $\mathcal{V}_i$ and $\mathcal{V}_o$, hence being of length $(D+L)$.
    Since both the input vectors $\textbf{x}_i$ and output vectors $\textbf{y}_i$ are highly sparse, this does not lead to any major computational burden in training.
\end{itemize}

It may be noted that label representation based solely on the input as considered by recent works~\cite{prabhu2018parabel, wydmuch2018no}, can be considered as a special case of our more general formulation of label representation. 
As also shown later in our empirical findings, in combination with shallow tree cascade of classifiers, partitioning of :

\begin{itemize}
\item output space representation ($\mathcal{V}_o$) yields competitive results compared to state-of-the-art classifiers in XMC such as \parabel.
\item joint representation ($\mathcal{V}_j$) further surpasses the state-of-the-art methods in terms of prediction performance and label diversity.
\end{itemize}


\subsection{Label partitioning via $K$-means clustering} 

Once we have obtained the representation $\vec{v}_{\ell}$ for each label $\ell$ in the set $\labelset = \set{1, \ldots, L}$, the next step is to iteratively partition $S$ into disjoint subsets. 
This is achieved by $K$-means clustering, which also presents many choices such as number of clusters and degree of balancedness among the clusters. 
Our goal, in this work, is to avoid propagation error in a deep tree cascade. We, therefore, choose a relatively large value of $K$ (e.g. $\ge 100$) which leads to shallow trees. 

The clustering step in \bonsai first partitions $\labelset$ into $\branchfactor$ disjoint sets $\set{\labelset_1, \ldots, \labelset_\branchfactor}$.
Each of the elements, $\labelset_k$, of the above set can be thought of as a meta-label which semantically groups actual labels together in one cluster.
Then, $\branchfactor$ child nodes of the root are created, each contains one of the partitions, $\set{\labelset_k}_{k=1}^\branchfactor$. 
The same process is repeated on each of the newly-created $\branchfactor$ child nodes in an iterative manner. 
In each sub-tree, the process terminates either when the node's depth exceeds pre-defined threshold $\treemaxdepth$ or the number of associated labels is no larger than $\branchfactor$, e.g, $\abs{S_k} \le \branchfactor$. 

Formally, without loss of generality, we assume a non-leaf node has labels $\set{1, \ldots, L}$. 
We aim at finding $\branchfactor$ cluster centers $\centerrepr_1, \ldots, \centerrepr_\branchfactor \in \mathbb{R}^{\eta}$, i.e., in an appropriate space (input, output, or joint) by optimizing the following :

\begin{equation}\label{eq:kmeans}
\min_{\centerrepr_1, \ldots, \centerrepr_\branchfactor \in \real^\eta}
\left[ \sum\limits_{k=1}^{\branchfactor}  \sum\limits_{\ell\in\centerrepr_i} \distfunc(\labelrepr_{\ell}, \centerrepr_k) \right]
\end{equation}
where $\distfunc(., .)$ represents a distance function and $\vect{v}_{\ell}$ represents the vector representation of the label ${\ell}$. 
The distance function is defined in terms of the dot product as follows : $\distfunc(\labelrepr_{\ell}, \centerrepr_k) = 1 - \labelrepr_{\ell}^T \cdot \centerrepr_k$.
The above problem is $\textbf{NP}$-hard and we use the standard $K$-means algorithm (also known as Lloyd's algorithm)~\cite{lloyd1982least}\footnote{We also tried $K\text{-means++}$ and observed that faster convergence did not out-weigh extra computation time for seed initialization.} for finding an approximate solution to equation (\ref{eq:kmeans}). 


\begin{figure*}[h!]
    \centering
    \begin{tabular}{cc}
        \includegraphics[width=0.48\textwidth]{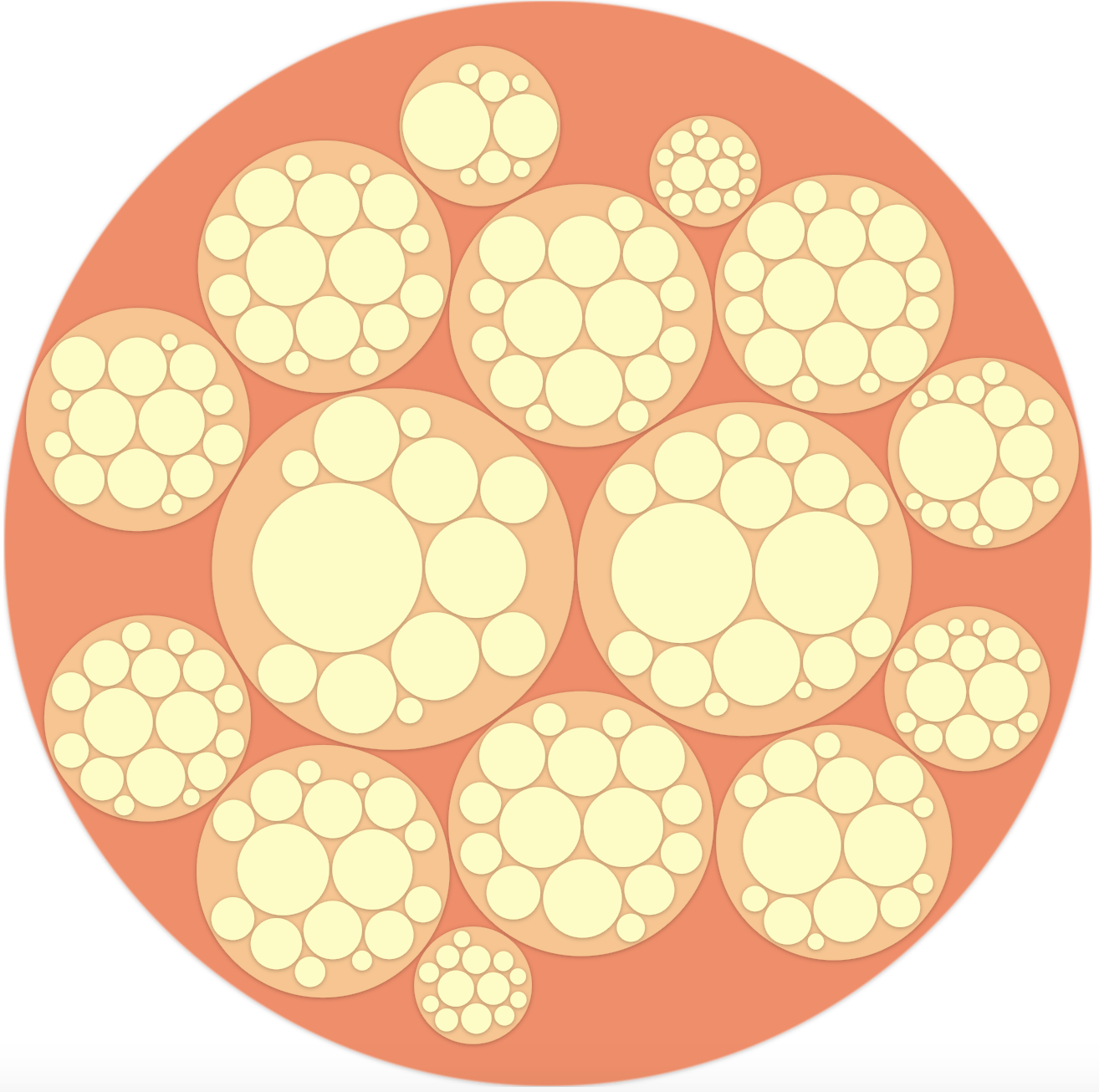}& \hfill \includegraphics[width=0.48\textwidth]{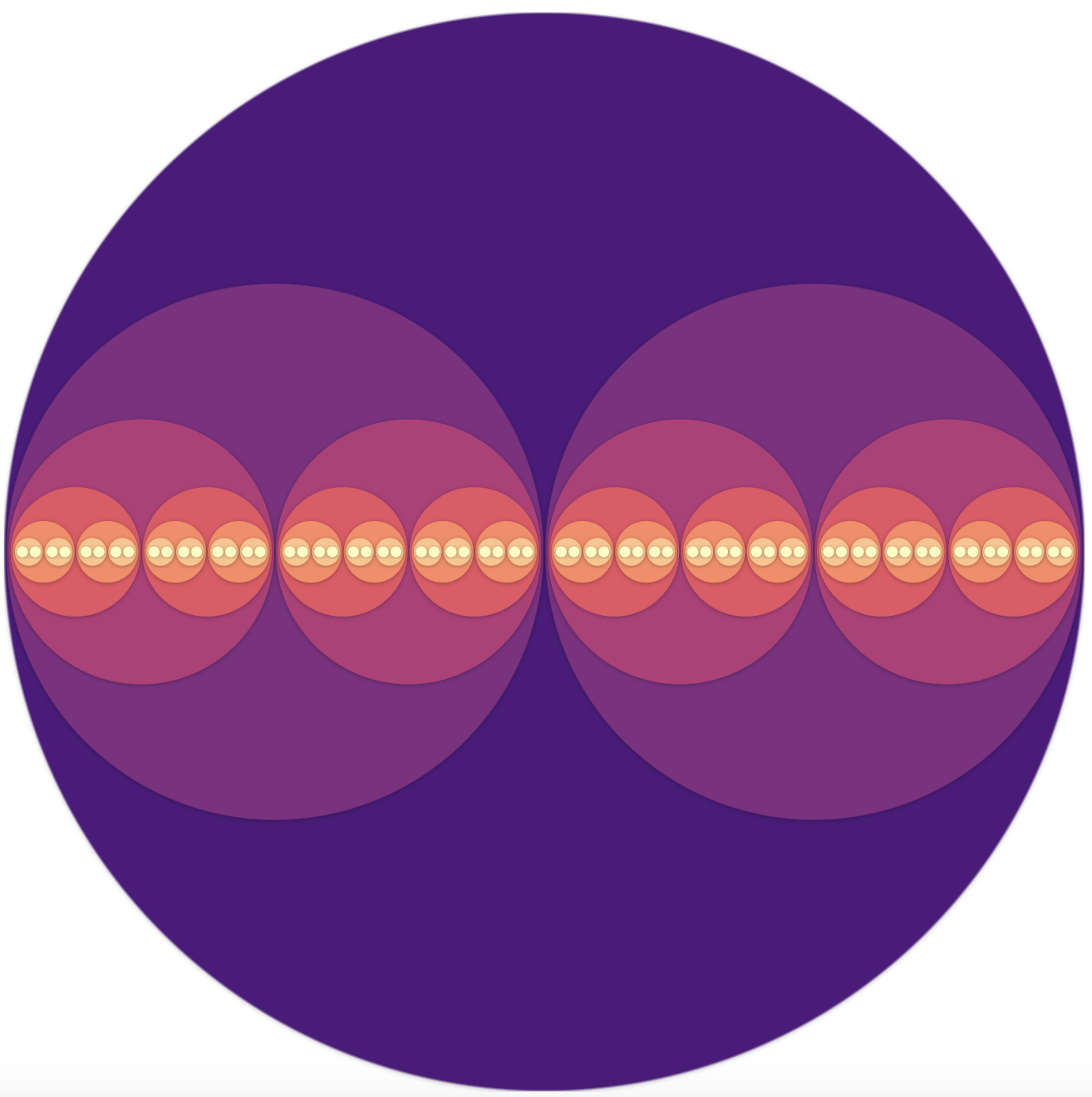} \\
        \bonsai $: \branchfactor=16$, tree depth 2 &      \parabel $: \branchfactor=2$, tree depth 6
    \end{tabular}        
    \caption{Comparison of partitioned label space by \bonsai and \parabel on \textbf{EURLex-4K} dataset. 
    Each circle corresponds to one label partition (also a tree node), the size of circle indicates the number of labels in that partition
    and lighter color indicates larger node level. 
    The largest circle is the whole label space.
    Note that \bonsai produces label partitions of varying sizes, while \parabel gives perfectly balanced partitioning.}
    \label{fig:diff}
\end{figure*}
The $K$-way unconstrained clustering in \bonsai has the following advantages over \parabel which enforces binary and balanced partitioning :
\begin{enumerate}
\item \textit{Initializing label diversity in partitioning :} By setting $\branchfactor > 2$, \bonsai allows a varied partitioning of the labels space, rather than grouping all labels in two clusters. This facet of \bonsai is especially favorable for tail labels by allowing them to be part of separate clusters if they are indeed very different from the rest of the labels. Depending on the similarity to other labels, each label can choose to be part of one of the $K$ clusters.

\item \textit{Sustaining label diversity :} \bonsai sustains the diversity in the label space by \textit{not enforcing} the balanced-ness constraint of the form, $\lvert|\labelset_k| - |\labelset_{k'}|\rvert \leq 1,  \forall 1 \leq k,k' \leq K$ (where $|.|$ operator is overloaded to mean set cardinality for the inner one and absolute value for the outer ones) among the partitions.
This makes the \bonsai partitions more data-dependent since smaller partitions with diverse tail-labels are very moderately penalized under this framework. 

\item \textit{Shallow tree cascade :} Furthermore, $K$-way unconstrained partitioning leads to shallower trees which are less prone propagation error in deeper trees constructed by \parabel. As we will show in Section~\ref{sec:experiment}, the diverse partitioning reinforced by shallower architecture leads to better prediction performance, and significant improvement is achieved on tail labels. 
\end{enumerate}

A pictorial description of the partitioning scheme of \bonsai and its difference compared to \parabel is also illustrated in Figure~\ref{fig:diff}.


\subsection{Learning node classifiers} 
\label{sec:learning-classifier}

Once the label space is partitioned into a diverse and shallow tree structure, we learn a $\branchfactor$-way \onevsall linear classifier at each node. 
These classifiers are trained independently using only the training examples that have at least one of the node labels. 
We distinguish the leaf nodes and non-leaf nodes in the following way :
(i) for non-leaf nodes, the classifier learns $\branchfactor$ linear classifiers separately, 
each maps to one of the $\branchfactor$ children. 
During prediction, the output of each classifier determines whether the test point should traverse down the corresponding child. 
(ii) for leaf nodes, the classifier learns to predict the actual labels on the node. 

Without loss of generality, given a node in the tree, denote by $\{c_k\}_{k=1}^{\branchfactor}$ as the set of its children. For the special case of leaf nodes, the set of children represent the final labels.
We learn $\branchfactor$ linear classifiers parameterized by $\set{\linearsep_1, \ldots, \linearsep_\branchfactor}$, where $\linearsep_k \in \real^{D}$ for $\forall k=1, \ldots, \branchfactor$. 
Each output label determines if the corresponding $\branchfactor$ children should be traversed or not.

For each of the child node $c_k$, 
we define the training data as $T_k = (\vect{X}_k, \vect{s}_k)$, 
where $\vect{X}_k=\set{\vect{x}_i \mid \vect{y}_{ik} = 1, i=1,\ldots,\trainsize}$. 
Let $\vect{s}_k \in \set{+1, -1}^\trainsize$ represent the vector of signs depending on whether $\vect{y}_{ik} = 1$ corresponds to +1 and $\vect{y}_{ik} = 0$ for -1.
We consider the following optimization problem for learning linear SVM with squared hinge loss and $\ell_2$-regularization 

\begin{equation}
    \min_{\linearsep_k} \left[ ||\linearsep_k||_2^2 + C \sum\limits_{i=1}^{|\vect{X}_k|} \mathcal{L}(s_{k_i} \linearsep_k^T \vect{x}_i)  \right]
\end{equation}
\begin{algorithm*}[h!]
  \SetKwInOut{Input}{Input}
  \SetKwInOut{Output}{Output}
  \Input{Training data $\instances=\set{(\pointx_i, \pointy_i)_{i=1}^N}$, where $\pointx_i \in \real^\dimension$ and $\pointy_i \in \{0,1\}^L$, branching factor $\branchfactor \ge 2$, maximum depth $\treemaxdepth$}
  \Output{a tree rooted at $\rootnode$}
  $\rootnode \defines \text{new node}$\;
  $\rootnode.\depth \defines 0$ \tcp*{$\depth$: node depth}  \
  $\rootnode.\labels \defines \set{1, \ldots, L}$  \tcp*{$\labels$: associated label set}\
  $\rootnode.\instances \defines \set{1, \ldots, N}$ \tcp*{$\instances$: associated training instance ids} \
  $\childrennodes \defines$ \grow($\rootnode$, $\treemaxdepth$, $\branchfactor$) \tcp*{grow the root recursively} \
  $\rootnode.\children \defines \childrennodes$ \tcp*{$\children:$ set of child nodes} \
  \Return $\rootnode$\;
  
  \SetKwFunction{FGrow}{\grow}
  \SetKwProg{Fn}{procedure}{:}{}
  \Fn{\FGrow{$\node$, $\treemaxdepth$, $\branchfactor$}}{
    $\labelparts \defines $$K$-means($\node.\labels$, $\branchfactor$) \tcp*{$\branchfactor$-way split of labels} \
    \For{$k = 1, \ldots, \branchfactor$}{
      $\childnode \defines$ new node\;
      $\childnode.\labels \defines S_k$\;
      $\childnode.\depth \defines \node.\depth + 1$ \;
      $\childnode.\instances \defines \set{i \in \node.\instances \mid \exists l \in S_k \text{ s.t. } \pointy_{il} = 1}$\;
      \eIf{$\branchfactor \ge \abs{\childnode.\labels}$ or $\childnode.\depth \ge \treemaxdepth$}{
        $\childnode.\linearsep \defines$ \solveonevsall($\childnode.\instances, \childnode.\labels$) \tcp*{$\childnode$ is a leaf}
      }
      {
        $\set{c_1, \ldots, c_\branchfactor} \defines$ \grow(\childnode, $\treemaxdepth$, $\branchfactor$) \tcp*{$\childnode$ is non-leaf} \
        $\childnode.\children \defines \set{c_1, \ldots, c_\branchfactor}$ \;
      }
      $\node.\linearsep \defines$ \solveonevsall($\node.\instances, \set{l_{\node_1}, \ldots, l_{\node_\branchfactor}}$)\tcp*{each $\childnode$ maps to a meta label}\
    }
    \Return $\set{\childnode}_{k=1}^\branchfactor$\;
  }
  \caption{Training algorithm: 
  \grow($\node, \branchfactor, \treemaxdepth$)\xspace partitions label space recursively and returns $\branchfactor$ children nodes of $\node$. 
  $K$-means($\labels, \branchfactor$) partitions label set $\labels$ into $\branchfactor$ disjoint sets using standard $K$-means algorithm. Label features are derived from training data $\instances$. 
  \solveonevsall($\instances, \set{l_1, \ldots, l_K}$) learns $K$ one-vs-rest linear classifiers $\set{\linearsep_k}_{k=1}^K$. 
  }
  \label{alg:train}
\end{algorithm*}
where $\mathcal{L}(z) = (\max(0,1-z))^2$.
This is solved using the Newton method based primal implementation in LIBLINEAR~\cite{fan2008liblinear}. To restrict the model size, and remove spurious parameters, thresholding of small weights is performed as in~\cite{dismec}.

The tree-structured architecture of \bonsai is illustrated in Figure~\ref{fig:arch}. The details of \bonsai's training procedure in the form of an algorithm is shown in Algorithm \ref{alg:train}. 
The partitioning process in Section \ref{sec:label-partitions} is described as the procedure \texttt{GROW} in the algorithm.
The \onevsall procedure is shown as \solveonevsall\xspace in Algorithm~\ref{alg:train}. 

\subsection{Prediction error propagation in shallow versus deep trees}
During prediction, a test point $\vect{x}$ traverses down the tree. 
At each non-leaf node, the classifier narrows down the search space by deciding which subset of child nodes $\vect{x}$ should further traverse. 
If the classifier decides not to traverse down some child node $c$, all descendants of $c$ will not be traversed. 
Later, as $\vect{x}$ reaches to one or more leaf nodes,
\onevsall classifiers are evaluated to assign probabilities to each label. 
\bonsai uses beam search to avoid the possibility of evaluating all nodes. 


The above search space pruning strategy implies errors made at non-leaf nodes could propagate to their descendants. 
\bonsai sets relatively large values to the branching factor $K$ (typically $100$), resulting in much shallower trees compared to \parabel, and hence significantly reducing error propagation, particularly for tail-labels.

%
%

More formally, given a data point $\pointx$ and a label $\ell$ that is relevant to $\pointx$, 
we denote $\leaf$ as the leaf node $\ell$ belongs to and $\ancestor(e)$ as the set of ancestor nodes of $\leaf$ and $\leaf$ itself.
Note that $\abs{\ancestor(e)}$ is path length from root to $\leaf$. 
Denote the parent of $n$ as  $\parent{n}$. 
We define the binary indicator variable $\event{n}$ to take value 1 if node $n$ is visited during prediction and 0 otherwise.
From the chain rule, the probability that $\ell$ is predicted as relevant for $\pointx$ is as follows:

\begin{align}
\predictproba &=\leafproba \\ \nonumber
       & \times \prod\limits_{n \in \ancestor(e)} \nodeproba \nonumber
\end{align}

Consider the Amazon-3M dataset with $L \approx 3 \times 10^6$, setting $\branchfactor=2$ produces a tree of depth 16. 
Assuming $\nodeproba=0.95$, for $\forall \node \in \parent{\node}$ and $\leafproba=1$, it gives $\predictproba = (0.95)^{16} \approx 0.46$. 
This is to say, even if $\nodeproba$ is high (e.g, 0.95) at each $n \in \ancestor(e)$, multiplying them together can result in small probability (e.g, 0.46) if the depth of the tree, i.e., $\abs{\ancestor(e)}$ is \textit{large}. 
We choose to mitigate this issue by increasing $\branchfactor$, and hence limiting the propagation error.

\section{Experimental Evaluation}
\label{sec:experiment}
In this section, we detail the dataset description, and the set up for comparison of the proposed approach against state-of-the-art methods in XMC.

\begin{table*}[ht!]
\centering
\resizebox{\textwidth}{!}{
\begin{tabular}{|c|c|c|c|c|c|c|}
\hline
  \textbf{Dataset} &  \# \textbf{Training } &  \# \textbf{Test } & \# \textbf{Labels} &  \# \textbf{Features}& \textbf{APpL} & \textbf{ALpP} \\
\hline
\textbf{EURLex-4K} &  15,539  & 3,809  & \textbf{3993} & 5000 & 25.7 & 5.3 \\
\textbf{Wikipedia-31K} &  14,146  & 6,616  & \textbf{30,938} & 101,938 & 8.5 & 18.6 \\
\textbf{WikiLSHTC-325K} &  1,778,351 & 587,084  & \textbf{325,056} & 1,617,899 & 17.4 & 3.2 \\
\textbf{Wikipedia-500K} & 1,813,391 & 783,743  &  \textbf{501,070} &  2,381,304 & 24.7 & 4.7 \\
\textbf{Amazon-670K} & 490,499 & 153,025 & \textbf{670,091}  & 135,909 & 3.9 & 5.4 \\
\textbf{Amazon-3M} & 1,717,899 & 742,507 & \textbf{2,812,281}  & 337,067 & 31.6 & 36.1 \\
\hline
\end{tabular}
}
\caption{\footnotesize Multi-label datasets used in the experiment. APpL and ALpP represent average points per label and average labels per point respectively.}
\label{tbl:datasets}
\end{table*}
\subsection{Dataset and evaluation metrics} 
We perform empirical evaluation on publicly available datasets from the XMC repository \footnote{\url{http://manikvarma.org/downloads/XC/XMLRepository.html}} curated from sources such as Amazon for item-to-item recommendation tasks and Wikipedia for tagging tasks.
The datasets of various scales in terms of number of labels are used, \textbf{EURLex-4K} consisting of approximately 4,000 labels to \textbf{Amazon-3M} consisting of 3 million labels.
The datasets also exhibit a wide range of properties in terms of number of training instances, features, and labels.
The detailed statistics of the datasets are shown in Table \ref{tbl:datasets}.

With applications in recommendation systems, ranking and web-advertising, the objective of the machine learning system in XMC is to correctly recommend/rank/advertise among the top-k slots.
We therefore use evaluation metrics which are standard and commonly used to compare various methods under the XMC setting - Precision@$k$ (\patk) and normalised Discounted Cumulative Gain (\ndcgatk).
Given a label space of dimensionality $L$, a predicted label vector $\hat{\textbf{y}} \in \mathbb{R}^L$ and a ground truth label vector $\textbf{y} \in \{0,1\}^L$ :
\begin{eqnarray}
\patk(\hat{\textbf{y}},\textbf{y}) & = &  \frac{1}{k} \sum_{\ell \in rank_k{(\hat{\textbf{y}})}}{\textbf{y}_\ell} \\
\ndcgatk(\hat{\textbf{y}},\textbf{y}) & = & \frac{DCG@k}{\sum_{\ell=1}^{\min(k, ||\textbf{y}||_0)}{\frac{1}{\log(\ell+1)}}} \label{eq:ndcg}
\end{eqnarray}
where $DCG@k = \frac{\textbf{y}_\ell}{\sum_{l=1}^{}{\frac{1}{\log(\ell+1)}}}$, and $rank_k(\hat{\textbf{y}})$ returns the $k$ largest indices of $\hat{\textbf{y}}$.

For better readability, we report the percentage version of above metrics (multiplying the original scores by 100). 
In addition, we consider $k \in \set{1,3,5}$.


\subsection{Methods for comparison} 

We consider three different variants of the proposed family of algorithms, \bonsai, which is based on the generalized label representations (discussed in Section \ref{sec:label-partitions}) combined with the shallow tree cascades. 
We refer the algorithms learnt by partitioning the input space, output space and the joint space as \bonsai-i, \bonsai-o, and \bonsai-io respectively.
These are compared against six state-of-the-art algorithms from each of the three main strands for XMC namely, label-embedding, tree-based and one-vs-all methods :
\begin{itemize}
  \item \textbf{Label-embedding methods}: Due to the fat-tailed distritbution of instances among labels, \texttt{SLEEC} \cite{bhatia2015sparse} makes a locally low-rank assumption on the label space, \texttt{RobustXML} \cite{xu2016robust} decomposes the label matrix into tail labels and non tail labels so as to enforce an embedding on the latter without the tail labels damaging the embedding. \texttt{LEML} \cite{yu2014large} makes a global low-rank assumption on the label space and performs a linear embedding on the label space. As a result, it gives much worse results, and is not compared explicitly in the interest of space.
  \item \textbf{Tree-based methods}: \texttt{FastXML}  \cite{prabhu2014fastxml} learns an ensemble of trees which partition the label space by directly optimizing an nDCG based ranking loss function, \texttt{PFastXML} \cite{Jain16} replaces the nDCG loss in \texttt{FastXML} by its propensity scored variant which is unbiased and assigns higher rewards for accurate tail label predictions, \parabel \cite{prabhu2018parabel} which has been described earlier in the paper. 
  \item \textbf{\onevsall methods}: \texttt{PD-Sparse} \cite{yen2016pd} enforces sparsity by exploiting the structure of a margin-maximizing loss with L1-penalty, \texttt{DiSMEC} \cite{dismec} learns one-vs-rest classifiers for every label with weight pruning to control model size.
\end{itemize}

\noindent Since we are considering only bag-of-words representation across all datasets, we do not compare against deep learning methods explicitly. However, it may be noted that despite using raw data and corresponding word-embeddings, deep learning methods in XMC are still sub-optimal in terms of prediction performance in XMC ~\cite{liu2017deep,joulin2017bag,kim2014convolutional}. More details on the performance of deep methods can be found in \cite{wydmuch2018no}.

\noindent \bonsai is implemented in C++ on a 64-bit Linux system. 
For all the datasets, we set the branching factor $\branchfactor=100$ at every tree depth. 
We will explore the effect of tree depth in details later. 
This results in depth-1 trees (excluding the leaves which represent the final labels) for smaller datasets such as \textbf{EURLex-4K}, \textbf{Wikipedia-31K} and depth-2 trees for larger datasets such as \textbf{WikiLSHTC-325K} and \textbf{Wikipedia-500K}.
\bonsai learns an ensemble of three trees similar to \parabel. 
For other approaches, the results were reproduced as suggested in the respective papers.

\begin{table*}[ht!]
\centering
\resizebox{\textwidth}{!}{
\begin{tabular}{|l||c|c|c||c|c||c|c||c|c|}
\hline
\multirow{2}{*}{Dataset} & \multicolumn{3}{c||}{Our Approach (\bonsai)} & \multicolumn{2}{c||}{Embedding~based} & \multicolumn{2}{c||}{Tree~based} &  \multicolumn{2}{c|}{Linear one-vs-rest}  \\
\cline{2-10}
&\bonsaii & \bonsaio & \bonsaiio & \texttt{SLEEC} & \texttt{RobustXML} &  \texttt{Fast-XML} & \parabel  & \texttt{PD-Sparse} & \texttt{DiSMEC} \\
\hline
\textbf{EURLex-4K} &&&&& & & & & \\
\textit{P@1} & \textbf{83.0} & 82.5 & 82.9 & 79.3 & 78.7 & 71.4 & 82.2 & 76.4 & 82.4\\
\textit{P@3} & \textbf{69.7} & 69.4 & 69.4 & 64.3 & 63.5 & 59.9 & 68.7 & 60.4 & 68.5\\
\textit{P@5} & \textbf{58.4} & 58.1 & 58.0 & 52.3 & 51.4 & 50.4 & 57.5 & 49.7 & 57.7\\



\textbf{Wikipedia-31K} &&&&& & & & & \\
\textit{P@1} & 84.7 & 84.70 & 84.8 & \textbf{85.5} & 85.5 & 82.5 & 84.2 & 73.8 &  84.1\\
\textit{P@3} & 73.6 & 73.57 & 73.6 & 73.6 & 74.0 & 66.6 & 72.5 & 60.9 & \textbf{74.6}\\
\textit{P@5} & 64.7 & 64.81 & 64.8 & 63.1 & 63.8 & 56.7 & 63.4 & 50.4 &  \textbf{65.9}\\ 


\textbf{WikiLSHTC-325K} &&&&& & & & & \\
\textit{P@1} & \textbf{66.6} & 63.4 & 65.8 & 55.5  & 53.5 & 49.3 & 65.0  & 58.2 &  64.4\\
\textit{P@3} & \textbf{44.5} & 42.8 & 44.1 & 33.8 & 31.8 & 32.7 & 43.2 & 36.3 & 42.5\\
\textit{P@5} & \textbf{33.0} & 32.0 & 32.7 & 24.0 & 29.9 & 24.0 & 32.0 & 28.7 & 31.5\\  

\textbf{Wikipedia-500K} &&&&& & & & & \\
\textit{P@1} & 69.2 & 68.7 & 69.1 & 48.2 & 41.3 & 54.1  & 68.7  & - &  \textbf{70.2}\\
\textit{P@3} & 49.8 & 48.8 & 49.7 & 29.4 &  30.1 & 35.5 & 49.6  & - & \textbf{50.6}\\
\textit{P@5} & 38.8 & 37.6 & 38.8 & 21.2 & 19.8 & 26.2 & 38.6 & - & \textbf{39.7}\\

\textbf{Amazon-670K} &&&&& & & & & \\
\textit{P@1} & 45.5 & 44.5 & \textbf{45.7} & 35.0 & 31.0 & 33.3  & 44.9  & -  &  44.7\\
\textit{P@3} & 40.3 & 39.8 & \textbf{40.6} & 31.2 & 28.0 & 29.3 & 39.8 & - & 39.7\\
\textit{P@5} & 36.5 & 36.4 & \textbf{36.9} & 28.5 & 24.0 &  26.1 & 36.0 & - & 36.1\\

\textbf{Amazon-3M} &&&&&& & &&\\
\textit{P@1} & 48.4 & 47.5 & \textbf{48.5} & - & - & 44.2  & 47.5  & - & 47.8\\
\textit{P@3}  & \textbf{45.6} & 44.7 & 45.5 & - & - & 40.8 &  44.6 & - & 44.9\\
\textit{P@5}  & 43.4 & 42.6 & \textbf{43.5} & - & - & 38.6  &  42.5 & - & 42.8\\

\hline
\end{tabular}
}
\caption{\patk (P@k) on benchmark datasets for $k=1,3\text{ and }5$. 
For each case of P@k and dataset, the best performed score is highlighted in bold. 
Entries marked "-" imply the corresponding method could not scale to the particular dataset, thus the scores are unavailable. }
\label{tbl:prec}
\end{table*}

\section{Experimental results} 
In this section, we report the main findings of our empirical evaluation.
\subsection{Precision@$k$} 
The comparison of \bonsai against other baselines is shown in Table \ref{tbl:prec}. The results are averaged over five runs with different initializations of the clustering algorithm. The important findings from these results are the following :
\begin{itemize}
	\item The competitive performance of the different variants of \bonsai shows the success and applicability of the notion of \textit{generalized label representation}, and their concrete realization discussed in section \ref{sec:label-partitions}. It further highlights that it is possible to enrich these representations further, and achieve better partitioning.
    \item The consistent improvement of \bonsai over \parabel on all datasets validates the choice of higher fanout and advantages of using \textit{shallow trees}.
    \item Another important insight from the above results is that when the average number of labels per training point are higher such as in Wikipedia-31K, Amazon-670K and Amazon-3M, the joint space label representation, used in \bonsaiio, leads to better partitioning and further improves the strong performance of input only label representation in \bonsaii.
    \item Even though \texttt{DiSMEC} performs slightly better on Wiki-500K and Wikipedia-31K, its computational complexity of training and prediction is orders of magnitude higher than \bonsai.
As a result, while \bonsai can be run in environments with limited computational resources, \texttt{DiSMEC} requires a distributed infrastructure for training and prediction.
\end{itemize}
\begin{figure*}[!h]
  \centering
  \begin{tabular}{ccc}
    \textbf{WikiLSHTC-325K} & \textbf{Wikipedia-500K} & \textbf{Amazon-3M} \\
    \begin{tikzpicture}[scale=0.55]
      \begin{axis}[
        major x tick style = transparent,
        height=7cm,
        ybar,
        enlarge x limits=0.25,
        symbolic x coords={\pwt{1},\pwt{3},\pwt{5}},
        xtick = data,
        ymin=15, ymax=40, 
        legend style={at={(0.03,0.97)},anchor=north west}]
        \addplot coordinates{ (\pwt{1}, 28.19) (\pwt{3}, 35.43) (\pwt{5}, 39.78)};
        \addplot +[postaction={pattern=north west lines}] coordinates{ (\pwt{1}, 26.76) (\pwt{3}, 33.27) (\pwt{5}, 37.34)};
        \addplot +[postaction={pattern=north east lines}] coordinates{ (\pwt{1}, 25.4) (\pwt{3}, 26.8) (\pwt{5}, 28.3)};
        \addplot coordinates{ (\pwt{1}, 16.35) (\pwt{3}, 20.99) (\pwt{5}, 23.56)};
        
        \legend{Bonsai, Parabel, PfastXML, FastXML}
      \end{axis}
    \end{tikzpicture}  &
                         \begin{tikzpicture}[scale=0.55]
                           \begin{axis}[
                             major x tick style = transparent,
                             height=7cm,
                             ybar,
                             enlarge x limits=0.25,
                             symbolic x coords={\pwt{1},\pwt{3},\pwt{5}},
                             xtick = data,
                             ymin=20, ymax=40, 
                             legend style={at={(0.03,0.97)},anchor=north west}]
                             \addplot coordinates{ (\pwt{1}, 27.80) (\pwt{3}, 32.77) (\pwt{5}, 36.26)};
                             \addplot +[postaction={pattern=north west lines}] coordinates{ (\pwt{1}, 26.75) (\pwt{3}, 31.60) (\pwt{5}, 34.92)};
                             \addplot +[postaction={pattern=north east lines}] coordinates{ (\pwt{1}, 22.2) (\pwt{3}, 21.3) (\pwt{5}, 21.6)};
                             \addplot coordinates{ (\pwt{1}, 22.5) (\pwt{3}, 21.8) (\pwt{5}, 22.4)};
                             
                           \end{axis}
                         \end{tikzpicture}&
                                            \begin{tikzpicture}[scale=0.55]
                                              \begin{axis}[
                                                major x tick style = transparent,
                                                height=7cm,
                                                ybar,
                                                enlarge x limits=0.25,
                                                symbolic x coords={\pwt{1},\pwt{3},\pwt{5}},
                                                xtick = data,
                                                ymin=0, ymax=30, 
                                                legend style={at={(0.03,0.97)},anchor=north west}]
                                                \addplot coordinates{ (\pwt{1}, 13.76) (\pwt{3}, 16.81) (\pwt{5}, 19.05)};
                                                \addplot +[postaction={pattern=north west lines}] coordinates{ (\pwt{1}, 12.81) (\pwt{3}, 15.63) (\pwt{5}, 17.76)};
                                                \addplot +[postaction={pattern=north east lines}] coordinates{ (\pwt{1}, 1) (\pwt{3}, 1) (\pwt{5}, 1)};
                                                \addplot coordinates{ (\pwt{1}, 9.77) (\pwt{3}, 11.69) (\pwt{5}, 13.25)};
                                                
                                              \end{axis}
                                            \end{tikzpicture} \\

    \begin{tikzpicture}[scale=0.55]
      \begin{axis}[
        major x tick style = transparent,
        height=7cm,
        ybar,
        enlarge x limits=0.25,
        symbolic x coords={\nwt{1},\nwt{3},\nwt{5}},
        xtick = data,
        ymin=15, ymax=40, 
        legend style={at={(0.03,0.97)},anchor=north west}]
        \addplot coordinates{ (\nwt{1}, 28.19) (\nwt{3}, 33.20) (\nwt{5}, 35.67)};
        \addplot +[postaction={pattern=north west lines}] coordinates{ (\nwt{1}, 26.76) (\nwt{3}, 31.26) (\nwt{5}, 33.57)};
        \addplot +[postaction={pattern=north east lines}] coordinates{ (\nwt{1}, 25.4) (\nwt{3}, 26.4) (\nwt{5}, 27.2)};
        \addplot coordinates{ (\nwt{1}, 16.35) (\nwt{3}, 19.56) (\nwt{5}, 21.02)};
        
      \end{axis}
    \end{tikzpicture} &
                        \begin{tikzpicture}[scale=0.55]
                          \begin{axis}[
                            major x tick style = transparent,
                            height=7cm,
                            ybar,
                            enlarge x limits=0.25,
                            symbolic x coords={\nwt{1},\nwt{3},\nwt{5}},
                            xtick = data,
                            ymin=20, ymax=40, 
                            legend style={at={(0.03,0.97)},anchor=north west}]
                            \addplot coordinates{ (\nwt{1}, 27.80) (\nwt{3}, 31.31) (\nwt{5}, 33.41)};
                            \addplot +[postaction={pattern=north west lines}] coordinates{ (\nwt{1}, 26.75) (\nwt{3}, 30.18) (\nwt{5}, 32.17)};
                            \addplot +[postaction={pattern=north east lines}] coordinates{ (\nwt{1}, 22.2) (\nwt{3}, 21.6) (\nwt{5}, 21.8)};
                            \addplot coordinates{ (\nwt{1}, 22.5) (\nwt{3}, 21.5) (\nwt{5}, 22.1)};
                            
                            \legend{Bonsai, Parabel, PfastXML, FastXML}
                          \end{axis}
                        \end{tikzpicture} & 
                                            \begin{tikzpicture}[scale=0.55]
                                              \begin{axis}[
                                                major x tick style = transparent,
                                                height=7cm,
                                                ybar,
                                                enlarge x limits=0.25,
                                                symbolic x coords={\nwt{1},\nwt{3},\nwt{5}},
                                                xtick = data,
                                                ymin=0, ymax=30, 
                                                legend style={at={(0.03,0.97)},anchor=north west}]
                                                \addplot coordinates{ (\nwt{1}, 13.78) (\nwt{3}, 16.05) (\nwt{5}, 17.62)};
                                                \addplot +[postaction={pattern=north west lines}] coordinates{ (\nwt{1}, 12.81) (\nwt{3}, 14.90) (\nwt{5}, 16.39)};
                                                \addplot +[postaction={pattern=north east lines}] coordinates{ (\nwt{1}, 1) (\nwt{3}, 1) (\nwt{5}, 1)};
                                                \addplot coordinates{ (\nwt{1}, 9.77) (\nwt{3}, 11.20) (\nwt{5}, 12.29)};
                                                
                                              \end{axis}
                                            \end{tikzpicture}
  \end{tabular}
  \caption{Comparison of $prec_{wt}@k$ (top row) and  $nDCG_{wt}@k$ (bottom row) over tree-based methods. 
    The reported metrics capture prediction performance over tail labels. Linear methods such as ProXML \cite{Babbar2019} and DiSMEC \cite{dismec} still remain the best on this metric.
}
  \label{fig:wt_plots_fig}
\end{figure*}
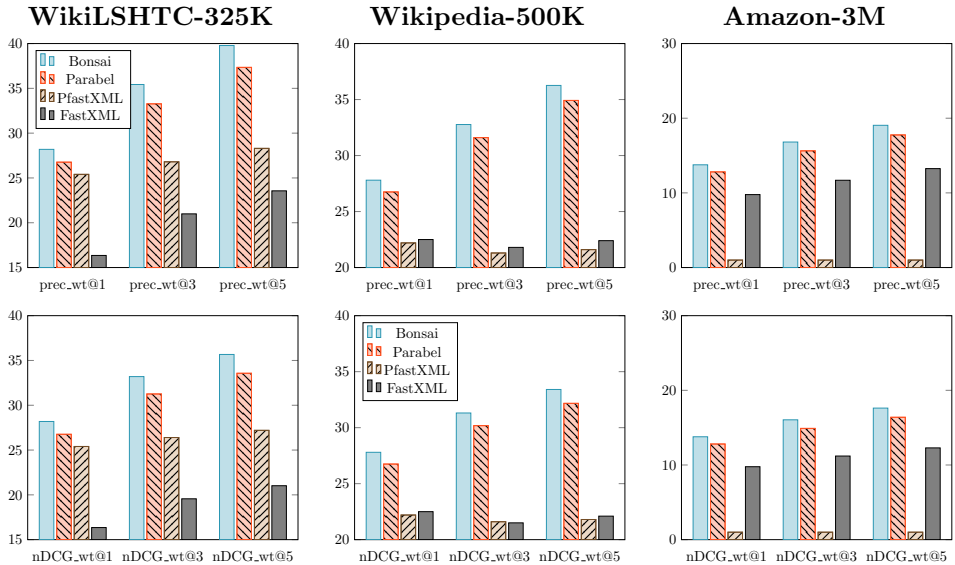

\subsection{Performance on tail labels}
We also evaluate prediction performance on tail labels using propensity scored variants of \patk and \ndcgatk. 
For label $\ell$, its propensity $p_{\ell}$ is related to number of its positive training instances $N_{\ell}$ by $p_{\ell}  \propto  1/\left(1+ e^{-\log(N_{\ell})}\right)$. With this formulation, $p_{\ell} \approx 1$ for head labels and $p_{\ell} \ll 1$ for tail labels. 
Let $\textbf{y} \in \{0,1\}^L$ and $\hat{\textbf{y}} \in \mathbb{R}^L$ denote the true and predicted label vectors respectively.
As detailed in \cite{Jain16}, propensity scored variants of $P@k$ and \ndcgatk are given by
\begin{eqnarray}
PSP@k(\hat{\textbf{y}},\textbf{y})  & := &  \frac{1}{k} \sum_{\ell \in rank_k{(\hat{\textbf{y}})}}{\textbf{y}_\ell}/p_\ell \label{eq:propp} \\
PSnDCG@k(\hat{\textbf{y}},\textbf{y}) \hspace{-0.1in} & := & \frac{PSDCG@k}{\sum_{\ell=1}^{\min(k, ||\textbf{y}||_0)}{\frac{1}{\log(\ell+1)}}}\hspace{-0.2in}
\label{eq:propndcg} \end{eqnarray}
where $PSDCG@k := \sum_{\ell \in rank_k{(\hat{\textbf{y}})}}{[\frac{\textbf{y}_\ell}{p_\ell\log(\ell+1)}]}$ , and $rank_k(\textbf{y})$ returns the $k$ largest indices of $\textbf{y}$.

To match against the ground truth, as suggested in \cite{Jain16}, we use $100 \cdot \mathbb{G}(\{\hat{\textbf{y}}\})/\mathbb{G}(\{\textbf{y}\})$ as the performance metric. For $M$ test samples, $\mathbb{G}(\{\hat{\textbf{y}}\}) = \frac{-1}{M}\sum_{i=1}^{M}\mathbb{L}(\hat{\textbf{y}}_i,\textbf{y})$, where $\mathbb{G}(.)$ and $\mathbb{L}(.,.)$ signify gain and loss respectively. 
The loss $\mathbb{L}(.,.)$ can take two forms, (i)$\mathbb{L}(\hat{\textbf{y}}_i,\textbf{y}) =  - PS\ndcgatk$, and (ii) $\mathbb{L}(\hat{\textbf{y}}_i,\textbf{y}) =  - PSP@k$.
This leads to the two metrics which are sensitive to tail labels and are denoted by $prec\_wt@k$, and $nDCG\_wt@k$.

Figure~\ref{fig:wt_plots_fig} shows the result w.r.t $prec\_wt@k$, and $nDCG\_wt@k$ among the tree-based approaches. 
Again, \bonsaii shows consistent improvement over \parabel. 
For instance, on WikiLSHTC-325K, the relative improvement over \parabel is approximately 6.7\% on $prec\_wt@5$. 
This further validates the applicability of the shallow tree architecture resulting from the design choices of $K$-way partitioning along with flexibility to allow unbalanced partitioning in \bonsai, 
which allows tail labels to be assigned into different partitions w.r.t the head ones.

\subsection{Unique label coverage}
\begin{table*}[ht!]
\centering
\begin{tabular}{|c|c|c|c|c|}
\hline
  \textbf{Dataset} & \textbf{Methods} & C@1 & C@3 & C@5\\
\hline
\multirow{ 2}{*}{\textbf{EUR-Lex}} & \parabel & \textbf{31.46}  & 43.11 & 54.38 \\
                                      & \bonsai & 31.38 & \textbf{44.09} & \textbf{55.61} \\ \hline
\multirow{ 2}{*}{\textbf{Wiki10}} & \parabel & 7.00  & 5.77 & 6.76 \\
                                  & \bonsai & \textbf{7.52} & \textbf{6.82} & \textbf{8.01} \\ \hline
\multirow{ 2}{*}{\textbf{WikiLSHTC}} & \parabel & 22.73  & 35.94 & 43.18 \\
                                  & \bonsai & \textbf{24.14} & \textbf{38.49} & \textbf{46.37} \\ \hline
\multirow{ 2}{*}{\textbf{Amazon-670k}} & \parabel & 32.73  & 33.77 & 38.82 \\
                                  & \bonsai & \textbf{33.28} & \textbf{34.76} & \textbf{40.11} \\ \hline
\multirow{ 2}{*}{\textbf{Amazon-3M}} & \parabel & 21.16  & 20.49 & 21.81 \\
                                  & \bonsai & \textbf{22.27} & \textbf{21.89} & \textbf{23.36} \\ \hline
\end{tabular}
\caption{
\footnotesize Coverage@k (C@k) statistics comparing \parabel and \bonsai. 
Along each C@k and dataset configuration, the best performing score is highlighted in bold. 
}
\label{tbl:coverage}
\end{table*}








We also evaluate \textit{coverage@k}, denoted $C@k$, which is the percentage of normalized unique labels present in an algorithm's top-$k$ labels. 
Let $\mathbf{P} = P_1 \cup P_2 \cup ... \cup P_M$ where $P_i = \{l_{i1}, l_{i2},.., l_{ik}\}$ i.e the set of top-$k$ labels predicted by the algorithm for test point $i$ and $M$ is the number of test points. Also, let $\mathbf{L} = L_1 \cup L_2 \cup ... \cup L_M$ where $L_i = \{g_{i1}, g_{i2},.., g_{ik}\}$ i.e the top-k propensity scored ground truth labels for test point $i$, then, coverage@k is given by  

\[
C@k = |\mathbf{P}| / |\mathbf{L}|
\]

The comparison between \bonsaii and \parabel of this metric on five different datasets is shown in Table \ref{tbl:coverage}. It shows that the proposed method is more effective in discovering correct unique labels. 
These results further reinforce the results in the previous section on the diversity preserving feature of \bonsai. 

\pgfplotsset{
    mystyle/.style={
        line width=3pt,mark size=5pt
    }
}

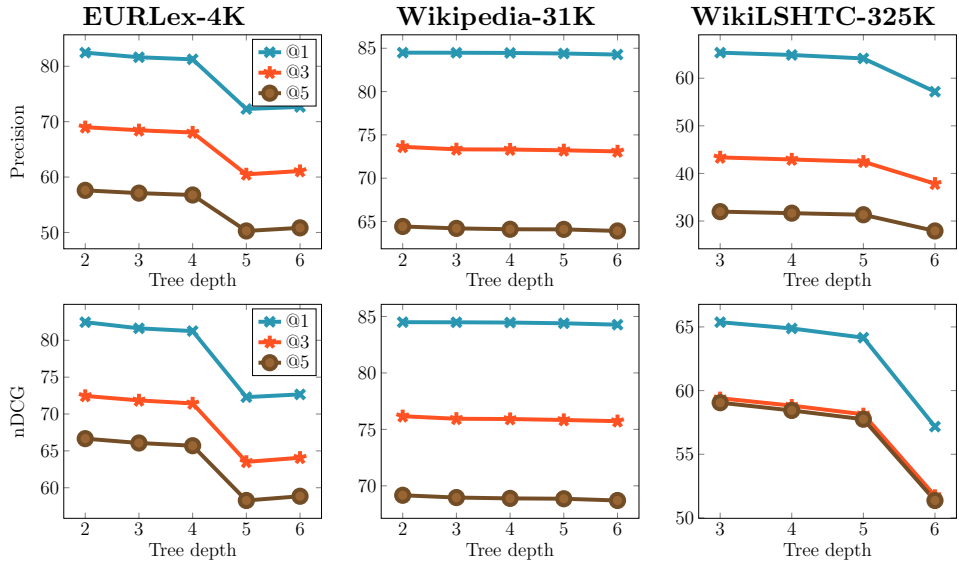
\begin{figure*}[h]
  \centering
  \begin{tabular}{ccc}
  \textbf{EURLex-4K} & \textbf{Wikipedia-31K} & \textbf{WikiLSHTC-325K} \\
  \begin{tikzpicture}[scale=0.5]
	\begin{axis}[
		xlabel=Tree depth,
		ylabel=Precision,
		every axis plot/.append style={ultra thick},
		ticklabel style = {font=\Large},
		label style={font=\Large},
        legend style={font=\Large}
	]
	\addplot +[ mark=x, mystyle] coordinates {
		(2, 82.46)
		(3, 81.627)
		(4, 81.25)
		(5, 72.29)
		(6, 72.662)
	};
	\addplot +[ mark=star, mystyle] coordinates {
		(2, 69.00)
		(3, 68.446)
		(4, 68.039)
		(5, 60.488)
		(6, 61.089)
	};
	\addplot +[ mark=*, mystyle] coordinates {
		(2, 57.6)
		(3, 57.10)
		(4, 56.76)
		(5, 50.27)
		(6, 50.83)
	};
	\legend{@1, @3, @5}
	\end{axis}
\end{tikzpicture} & \begin{tikzpicture}[scale=0.5]
	\begin{axis}[
		xlabel=Tree depth,
		ticklabel style = {font=\Large},
		every axis plot/.append style={ultra thick},
		label style={font=\Large},
		legend style={font=\Large}
	]
	\addplot +[mark=x, mystyle] coordinates {
		(2, 84.494)
		(3, 84.484)
		(4, 84.461)
		(5, 84.396)
		(6, 84.273)
	};
	\addplot +[mark=star, mystyle] coordinates {
		(2, 73.622)
		(3, 73.329)
		(4, 73.308)
		(5, 73.22)
		(6, 73.092)
	};
	\addplot +[mark=*, mystyle] coordinates {
		(2, 64.428)
		(3, 64.205)
		(4, 64.104)
		(5, 64.092)
		(6, 63.909)
	};
	\end{axis}
\end{tikzpicture} & \begin{tikzpicture}[scale=0.5]
	\begin{axis}[
		xlabel=Tree depth,
		xtick={3,4,5,6},
		ticklabel style = {font=\Large},
		every axis plot/.append style={ultra thick},
		label style={font=\Large}]
	\addplot +[ mark=x, mystyle] coordinates {
		(3, 65.373)
		(4, 64.881)
		(5, 64.156)
		(6, 57.174)
	};
	\addplot +[ mark=star, mystyle] coordinates {
		(3, 43.36)
		(4, 42.93)
		(5, 42.46)
		(6, 37.84)
	};
	\addplot +[ mark=*, mystyle] coordinates {
		(3, 31.97)
		(4, 31.65)
		(5, 31.317)
		(6, 27.913)
	};
	\end{axis}
\end{tikzpicture} \\
    
    \begin{tikzpicture}[scale=0.5]
	\begin{axis}[
		xlabel=Tree depth,
		ylabel=nDCG,
		every axis plot/.append style={ultra thick},
		ticklabel style = {font=\Large},
		label style={font=\Large},
		legend style={font=\Large}]
	\addplot +[ mark=x, mystyle] coordinates {
		(2, 82.459)
		(3, 81.627)
		(4, 81.25)
		(5, 72.29)
		(6, 72.662)
	};
	\addplot+[ mark=star, mystyle] coordinates {
		(2, 72.453)
		(3, 71.849)
		(4, 71.435)
		(5, 63.51)
		(6, 64.07)
	};
	\addplot+[ mark=*, mystyle] coordinates {
		(2, 66.66)
		(3, 66.07)
		(4, 65.70)
		(5, 58.27)
		(6, 58.845)
	};
	\legend{@1, @3, @5}
	\end{axis}
\end{tikzpicture} & \begin{tikzpicture}[scale=0.5]
	\begin{axis}[
		xlabel=Tree depth,
		every axis plot/.append style={ultra thick},
		ticklabel style = {font=\Large},
		label style={font=\Large},]
	\addplot +[ mark=x, mystyle] coordinates {
		(2, 84.494)
		(3, 84.484)
		(4, 84.461)
		(5, 84.396)
		(6, 84.273)
	};
	\addplot +[ mark=star, mystyle] coordinates {
		(2, 76.165)
		(3, 75.939)
		(4, 75.917)
		(5, 75.831)
		(6, 75.726)
	};
	\addplot +[mark=*, mystyle] coordinates {
		(2, 69.164)
		(3, 68.962)
		(4, 68.888)
		(5, 68.857)
		(6, 68.702)
	};
	\end{axis}
\end{tikzpicture} & \begin{tikzpicture}[scale=0.5]
	\begin{axis}[
		xlabel=Tree depth,
		xtick={3,4,5,6},
		every axis plot/.append style={ultra thick},
		ticklabel style = {font=\Large},
		label style={font=\Large}]
	\addplot +[mark=x, mystyle] coordinates {
		(3, 65.373)
		(4, 64.881)
		(5, 64.156)
		(6, 57.174)
	};
	\addplot +[mark=star, mystyle] coordinates {
		(3, 59.4)
		(4, 58.82)
		(5, 58.14)
		(6, 51.73)
	};
	\addplot +[mark=*, mystyle] coordinates {
		(3, 59.05)
		(4, 58.44)
		(5, 57.75)
		(6, 51.37)
	};
	\end{axis}
\end{tikzpicture} 
  \end{tabular}
  \caption{Effect of tree depth: \bonsai trees with different depths are evaluated w.r.t \patk (top row) and \ndcgatk (bottom row). 
  As tree depth increases, performance tends to drop. }
  \label{fig:depth-exp-fig}
\end{figure*}

\subsection{Impact of tree depth} We next evaluate prediction performance produced by \bonsai trees with different depth values. 
We set the fan-out parameter $\branchfactor$ appropriately to achieve the desired tree depth. 
For example, to partition 4,000 labels into a hierarchy of depth two, we set $\branchfactor=64$. 

In Figure~\ref{fig:depth-exp-fig}, we report the result on three datasets, averaged over ten runs under each setting. 
The trend is consistent - as the tree depth increases, prediction accuracy tends to drop, though it is not very stark for Wikipedia-31K. 

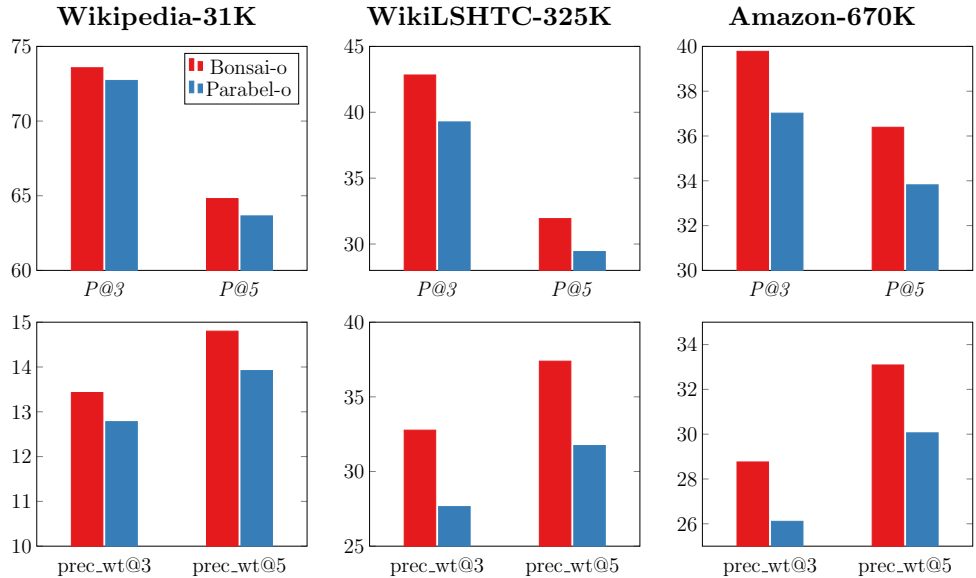
\begin{figure*}[!h]
  \centering
  \begin{tabular}{ccc}
    \textbf{Wikipedia-31K} & \textbf{WikiLSHTC-325K} & \textbf{Amazon-670K} \\
    \begin{tikzpicture}[scale=0.55]
      \begin{axis}[
        major x tick style = transparent,
        height=7cm,
        bar width=0.3in,
        ybar,
        cycle list/Set1,
        every axis plot/.append style={fill},
        enlarge x limits=0.5,
        symbolic x coords={\textit{P@3},\textit{P@5}},
        xtick = data,
        ymin=60, ymax=75, 
        ticklabel style = {font=\Large},
        legend style={at={(0.97,0.97)},anchor=north east, font=\Large}]
        \addplot coordinates{ (\textit{P@3}, 73.57) (\textit{P@5}, 64.81) };
        \addplot coordinates{ (\textit{P@3}, 72.72) (\textit{P@5}, 63.65) };
        
        \legend{Bonsai-o, Parabel-o}
      \end{axis}
    \end{tikzpicture}  &
                         \begin{tikzpicture}[scale=0.55]
                           \begin{axis}[
                             major x tick style = transparent,
                             height=7cm,
                             bar width=0.3in,
                             ybar,
                              cycle list/Set1,
                              every axis plot/.append style={fill},
                             enlarge x limits=0.5,
                             symbolic x coords={\textit{P@3},\textit{P@5}},
                             xtick = data,
                             ymin=28, ymax=45, 
                             ticklabel style = {font=\Large},
                             legend style={at={(0.03,0.97)},anchor=north west}]
                             \addplot coordinates{ (\textit{P@3}, 42.83) (\textit{P@5}, 31.93) };
                             \addplot coordinates{ (\textit{P@3}, 39.27) (\textit{P@5}, 29.43) };
                             
                           \end{axis}
                         \end{tikzpicture}&
                                            \begin{tikzpicture}[scale=0.55]
                                              \begin{axis}[
                                                major x tick style = transparent,
                                                height=7cm,
                                                bar width=0.3in,
                                                ybar,
                                                cycle list/Set1,
                                                every axis plot/.append style={fill},
                                                enlarge x limits=0.5,
                                                symbolic x coords={\textit{P@3},\textit{P@5}},
                                                xtick = data,
                                                ymin=30, ymax=40,
                                                ticklabel style = {font=\Large},
                                                legend style={at={(0.03,0.97)},anchor=north west}]
                                                \addplot coordinates{ (\textit{P@3}, 39.78) (\textit{P@5}, 36.39) };
                                                \addplot coordinates{ (\textit{P@3}, 37.02) (\textit{P@5}, 33.82) };
                                                
                                              \end{axis}
                                            \end{tikzpicture} \\
        \begin{tikzpicture}[scale=0.55]
          \begin{axis}[
            major x tick style = transparent,
            height=7cm,
            bar width=0.3in,
            ybar,
            cycle list/Set1,
            every axis plot/.append style={fill},
            enlarge x limits=0.5,
            symbolic x coords={\pwt{3},\pwt{5}},
            xtick = data,
            ymin=10, ymax=15, 
            ticklabel style = {font=\Large},
            legend style={at={(0.97,0.97)},anchor=north east, font=\Large}]
            \addplot coordinates{ (\pwt{3}, 13.43) (\pwt{5}, 14.80) };
            \addplot coordinates{ (\pwt{3}, 12.78) (\pwt{5}, 13.92) };
            
          \end{axis}
        \end{tikzpicture}  &
                             \begin{tikzpicture}[scale=0.55]
                               \begin{axis}[
                                 major x tick style = transparent,
                                 height=7cm,
                                 bar width=0.3in,
                                 ybar,
                                  cycle list/Set1,
                                  every axis plot/.append style={fill},
                                 enlarge x limits=0.5,
                                 symbolic x coords={\pwt{3},\pwt{5}},
                                 xtick = data,
                                 ymin=25, ymax=40, 
                                 ticklabel style = {font=\Large},
                                 legend style={at={(0.03,0.97)},anchor=north west}]
                                 \addplot coordinates{ (\pwt{3}, 32.77) (\pwt{5}, 37.39) };
                                 \addplot coordinates{ (\pwt{3}, 27.65) (\pwt{5}, 31.74) };
                                 
                               \end{axis}
                             \end{tikzpicture}&
                                                \begin{tikzpicture}[scale=0.55]
                                                  \begin{axis}[
                                                    major x tick style = transparent,
                                                    height=7cm,
                                                    bar width=0.3in,
                                                    ybar,
                                                    cycle list/Set1,
                                                    every axis plot/.append style={fill},
                                                    enlarge x limits=0.5,
                                                    symbolic x coords={\pwt{3},\pwt{5}},
                                                    xtick = data,
                                                    ymin=25, ymax=35,
                                                    ticklabel style = {font=\Large},
                                                    legend style={at={(0.03,0.97)},anchor=north west}]
                                                    \addplot coordinates{ (\pwt{3}, 28.76) (\pwt{5}, 33.09) };
                                                    \addplot coordinates{ (\pwt{3}, 26.11) (\pwt{5}, 30.06) };
                                                    
                                                  \end{axis}
                                                \end{tikzpicture} \\
  \end{tabular}
  \caption{Comparison of \patk and \pwt{k} scores of Bonsai-o and Parabel-o over three benchmark datasets.}
  \label{fig:only-output-exp-fig}
\end{figure*}

Furthermore, in Figure~\ref{fig:only-output-exp-fig}, we show that the shallow architecture is an integral part of the success of the \bonsai family of algorithms. To demonstrate this, we plugged in the label representation used in \bonsaio into \parabel, called \texttt{Parabel-o} in the figure.  As can be seen, \bonsaio outperforms \texttt{Parabel-o} by a large margin showing that shallow trees substantially alleviate the prediction error.


\subsection{Training and prediction time} Growing shallower trees in \bonsai comes at a slight price in terms of training time. It was observed that \bonsai leads to approximately 2-3x increase in training time compared to \parabel. For instance, on three cores, \parabel take one hour for training on WikiLSHTC-325K dataset, while \bonsai takes approximately three hours for the same task. 
However, it may also be noted that the training process can be performed in an offline manner. Though, unlike \parabel, \bonsai does not come with logarithmic dependence on the number of labels for the computational complexity of prediction. However, its prediction time is typically in milli-seconds, and hence it remains quite practical in XMC applications with real-time constraints such as recommendation systems and advertising.
\section{Conclusion}
\label{sec:conclusion}
In this paper, we present \bonsai, which is a class of algorithms for learning shallow trees for label partitioning in extreme multi-label classification. 
Compared to the existing tree-based methods, it improves this process in two fundamental ways. 
\textit{Firstly}, it generalizes the notion of label representation beyond the input space representation, and shows the efficacy of output space representation based on its co-occurrence with other labels, and by further combining these in a joint representation. 
\textit{Secondly}, by learning shallow trees which prevent error propagation in the tree cascade and hence improving the prediction accuracy and tail-label coverage. 
The synergizing effects of these two ingredients enables \bonsai to retain the training speed comparable to tree-based methods, while achieving better prediction accuracy as well as significantly better tail-label coverage. 
As a future work, the generalized label representation can be further enriched by combining with embeddings from raw text. This can lead to the amalgamation of methods studied in this paper with those that are based on deep learning.

%
%
%
%
%
%
%
\bibliographystyle{splncs04}
\bibliography{biblio-new}
\end{document}